\documentclass{article}

\usepackage[final, nonatbib]{nips_2016}

\usepackage{hyperref}
\usepackage[utf8]{inputenc} 
\usepackage[T1]{fontenc}    
\usepackage{url}            
\usepackage{booktabs}       
\usepackage{amsfonts}       
\usepackage{nicefrac}       
\usepackage{microtype}      
\usepackage{graphicx}
\usepackage{afterpage}
\usepackage{float}

\title{Visualizing Residual Networks}

\author{
  Brian Chu \\
  University of California, Berkeley\\
  \texttt{brian.c@berkeley.edu} \\
  \And
  Daylen Yang \\
  University of California, Berkeley\\
  \texttt{daylen@berkeley.edu} \\
  \And
  Ravi Tadinada \\
  University of California, Berkeley\\
  \texttt{rtadinada@berkeley.edu} \\
}

\begin{document}

\maketitle

\begin{abstract}
Residual networks are the current state of the art on ImageNet. Similar work in the direction of utilizing shortcut connections has been done extremely recently with derivatives of residual networks and with highway networks. This work potentially challenges our understanding that CNNs learn layers of local features that are followed by increasingly global features. Through qualitative visualization and empirical analysis, we explore the purpose that residual skip connections serve. Our assessments show that the residual shortcut connections force layers to refine features, as expected. We also provide alternate visualizations that confirm that residual networks learn what is already intuitively known about CNNs in general.
\end{abstract}

\section{Introduction}

In 2015, deep residual networks won 1st place on the ILSVRC classification task \cite{He15}. We seek to understand the qualitative characteristics underlying the shortcut connections and identity mappings that motivated He et al. towards this network architecture. To this end, we implement two visualizations of the feature maps after each residual building block: the top-9 images that maximally activate a unit in a given channel, and a guided backpropagation \cite{Springenberg14} visualization that unit's activation.

From these visualizations, we can visually confirm He et al.'s intuition that preconditioning layers to the identity mapping helps, and that it is easier to learn functions relative to the identity mapping. Specifically, we observe that residual layers of the same dimensionality learn features that get refined and sharpened.

\subsection{Related Work}

Zeiler et al. performed similar visualizations of AlexNet features \cite{Zeiler14,Krizhevsky12}, introducing a deconvolutional transformation that consists of taking the desired unit activation that one wishes to visualize and moving backwards through a series of deconvolutional steps. Instead of mapping from pixel space to feature space, deconvolution maps from feature space to pixel space. To move backwards through max-pooling, an unpooling step is performed, where the units that were selected as the maximal units in the forward pass are assigned the values being propagated backwards. To reverse convolution, a transposed convolution (also known as a fractionally strided convolution \cite{Dumoulin16}) is performed, using the same parameters as the network's learned convolutional layer. Finally, moving backwards through rectification was performed with rectification of the backwards inputs. This approach constructs in pixel space a visualization of the parts of an image contributing most to the activation of a given unit.

Springenberg et al. built upon the deconvolutional approach, developing \textit{guided backpropagation} \cite{Springenberg14}. Guided backpropagation is a tweak of the deconvolutional approach, where units that are rectified to zero in the forward pass (because they have a negative value) are also set to zero in the deconvolutional pass. This was demonstrated to be visually superior to visualization based on deconvolution, for the networks that Springenberg et al. examined.

Finally, Yosinski et al. visualized AlexNet using a variety of approaches, including the previous approaches and optimization to synthetically generate maximally-activating images \cite{Yosinski15}.

\section{Experiments and Architecture}

\begin{figure}
  \centering
  \includegraphics[width=3.0cm]{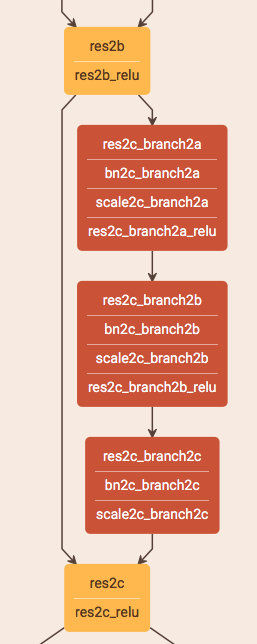}
  \includegraphics[width=3.0cm]{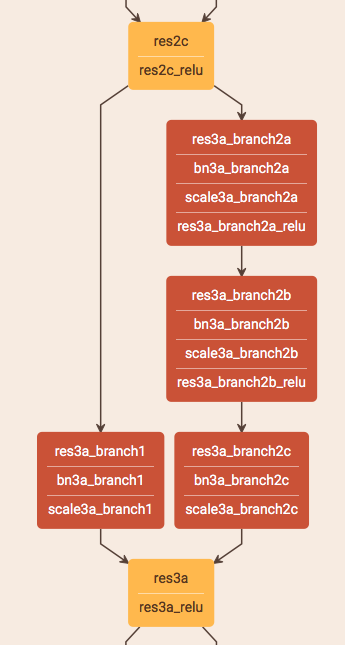}
  \caption{At left: basic shortcut block. At right: projection shortcut block.}
  \label{fig:blocks}
\end{figure}

The CNN architecture that we used was a pretrained 50-layer residual network \cite{He15}. A visualization of the architecture can be found online (\url{http://ethereon.github.io/netscope/#/gist/db945b393d40bfa26006}) \cite{Netscope}. This residual architecture consists of a single convolution layer (conv1), then a max-pooling layer, then a series of residual shortcut building blocks. There are two flavors of residual shortcut block, both shown in 
figure~\ref{fig:blocks}. The first flavor consists of 1x1, 3x3, and 1x1 convolutional layers, with a shortcut adding the inputs to the 1x1 convolution to the outputs of the final 1x1 convolution, performed elementwise. This conditions the stack of 3 layers towards the identity mapping.

The second flavor is called a projection shortcut. It consists of the same stack of convolutional layers, except the shortcut now contains a single 1x1 convolution.

The 50-layer residual architecture consists of an initial preojection block (2a), followed by two basic blocks (2b, 2c). This is followed by a projection block (3a), and a series of basic blocks (3b, 3c). This basic pattern is repeated twice more (4a, then 4b, 4c, 4d, 4e, 4f. Also: 5a then 5b, 5c).

Two more differences are associated with the projection shortcut: reduced spatial dimension due to a stride of 2, and an increase in the number of channels. This means that building blocks named with the same number (e.g. 2a, 2b and 2c) contain the same number of output channels.

For our visualizations, we used code from Yosinski et al., modifying it to work with residual networks and modifying it to implement guided backpropagation instead of deconvolution \cite{Yosinski15}.

To obtain the top-9 images maximally activating each unit, we used the ImageNet validation set. The units that are visualized are those with the highest activation values, across spatial dimensions, and across all images in the dataset. This means the visualization for a given filter is that of a specific spatial unit (e.g. position [20, 31] in the feature map for the channel).

\begin{figure}[H]
  \centering
  \label{fig:res2a}
  \includegraphics[width=6.9cm]{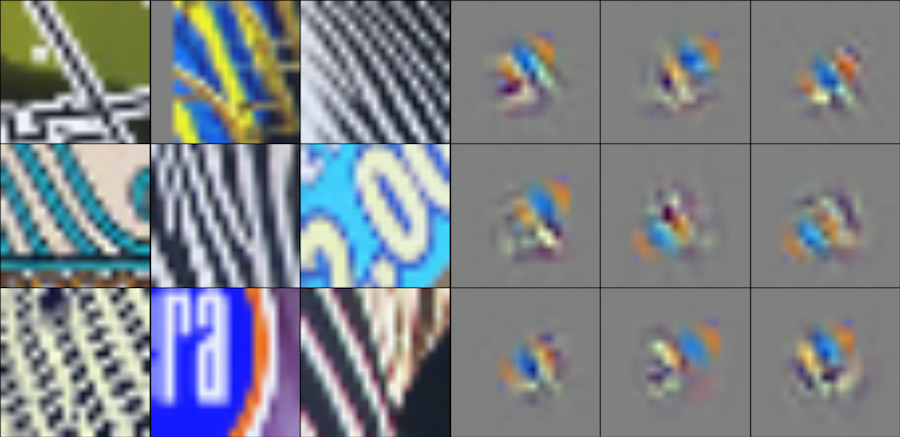}
  \includegraphics[width=6.9cm]{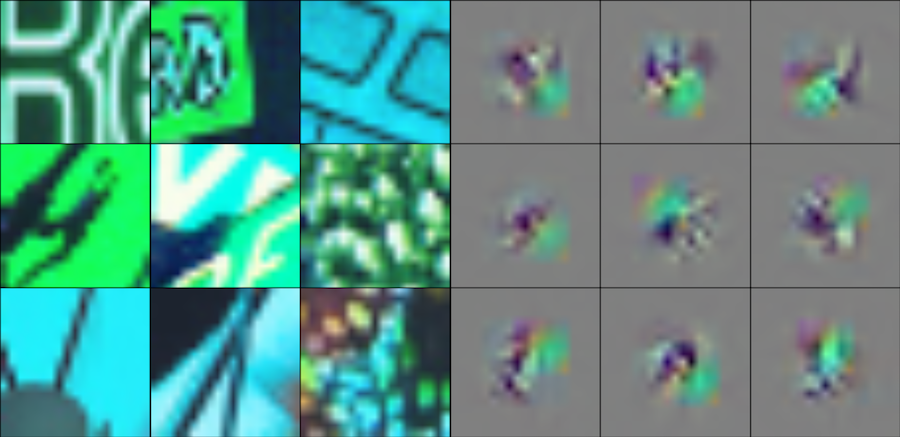}
  \includegraphics[width=6.9cm]{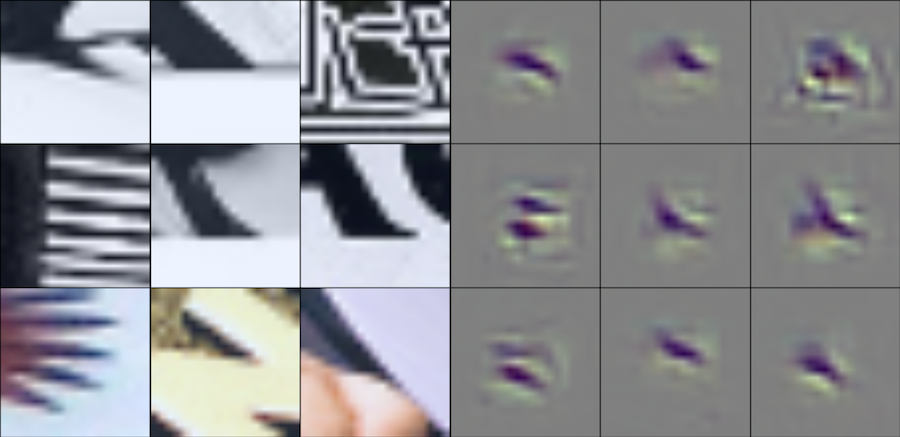}
  \includegraphics[width=6.9cm]{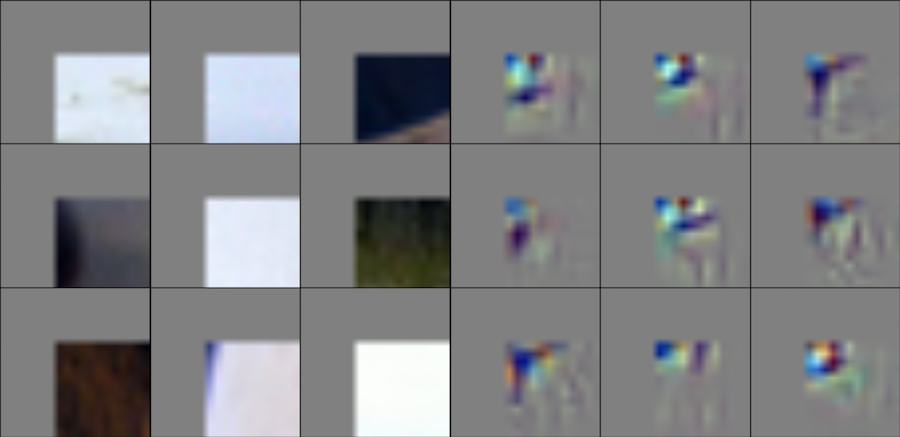}
  \caption{Res2a feature visualizations (randomly chosen). From left to right, top to bottom: channels 12, 79, 150, and 210. For each channel, at left are top-9 image patches. At right are corresponding guided backpropagation visualizations. The large gray borders are due to differing receptive field sizes (due to padding).}

  \includegraphics[width=6.9cm]{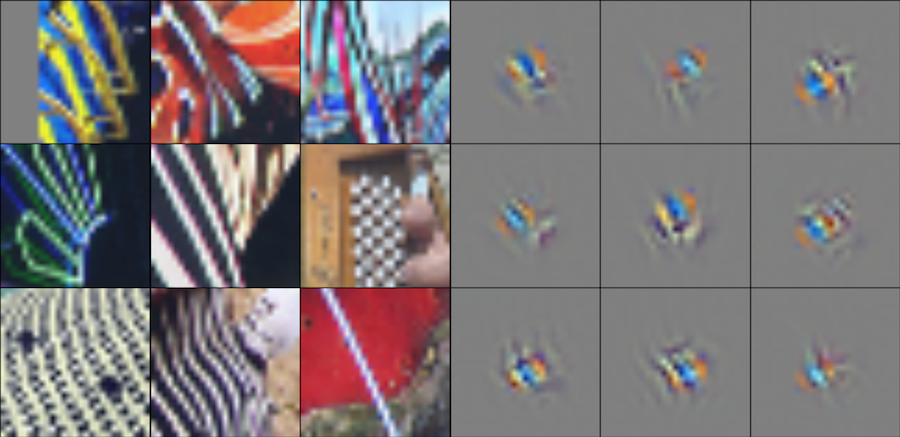}
  \includegraphics[width=6.9cm]{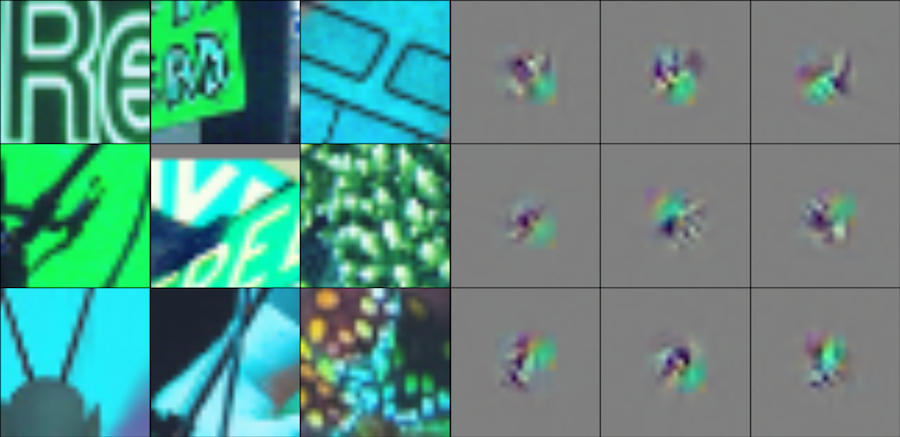}
  \includegraphics[width=6.9cm]{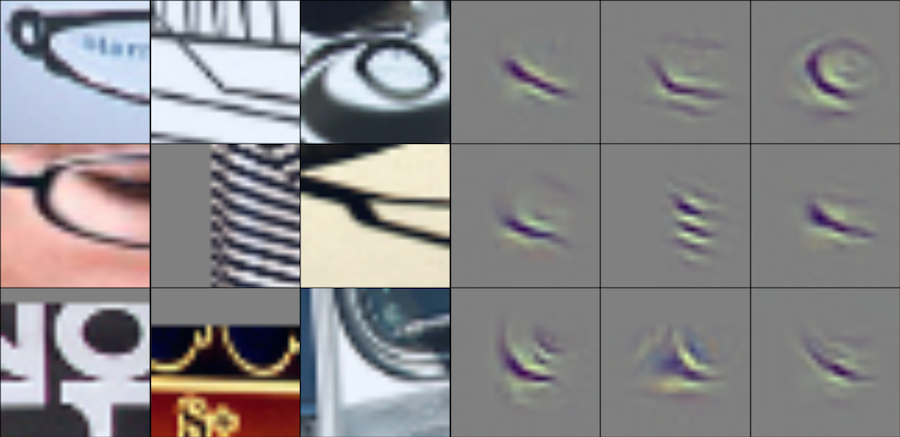}
  \includegraphics[width=6.9cm]{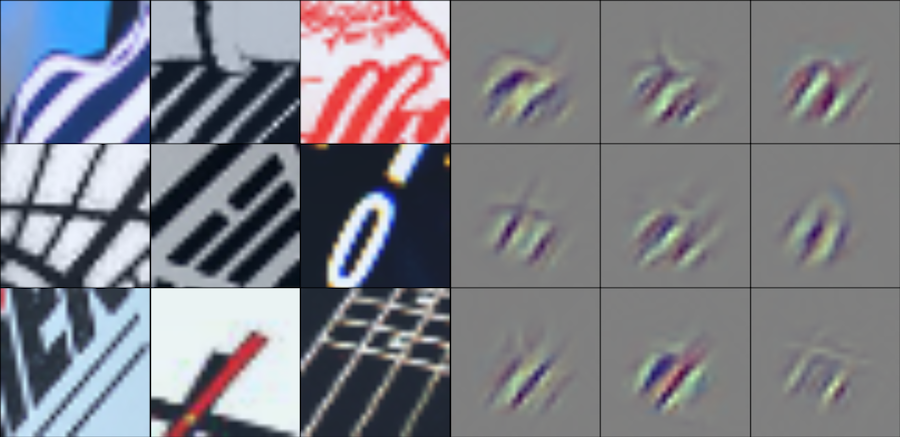}
  \noindent\makebox[\linewidth]{\rule{6.9cm}{0pt}}
  \includegraphics[width=6.9cm]{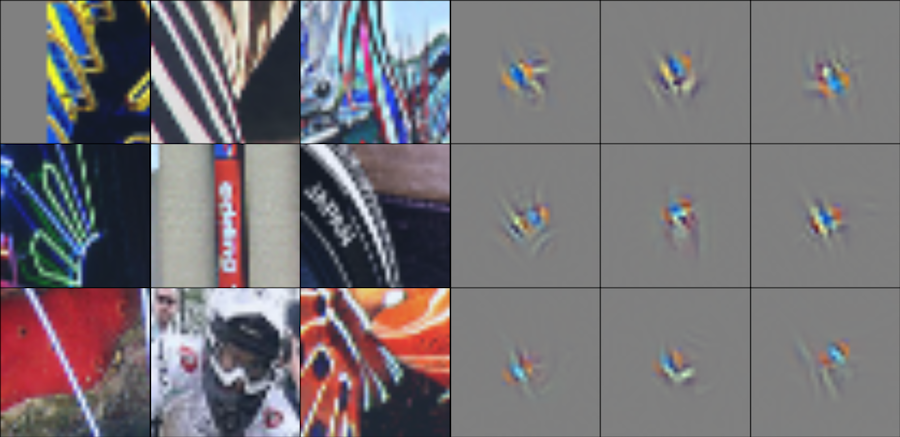}
  \includegraphics[width=6.9cm]{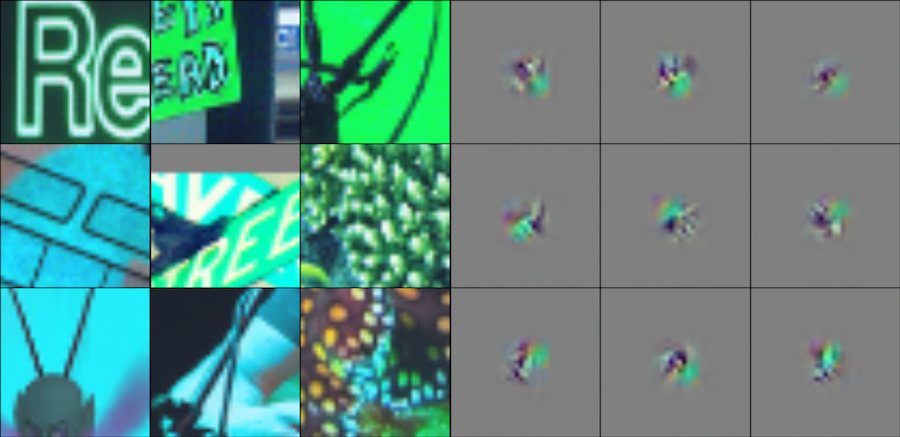}
  \includegraphics[width=6.9cm]{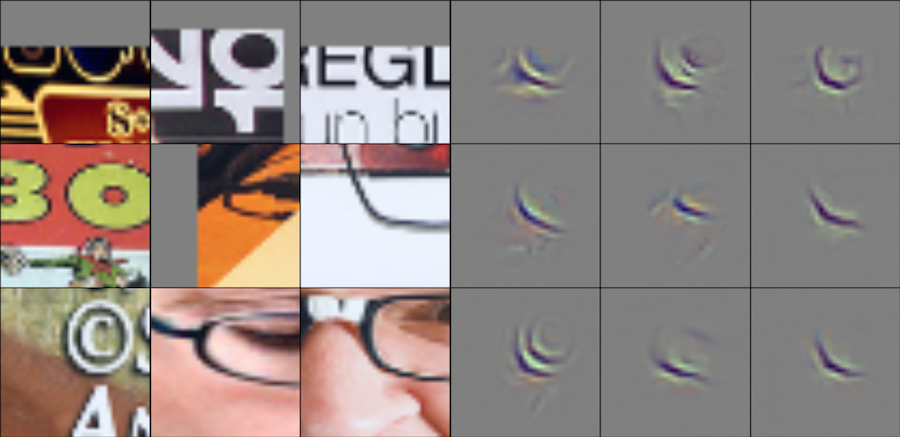}
  \includegraphics[width=6.9cm]{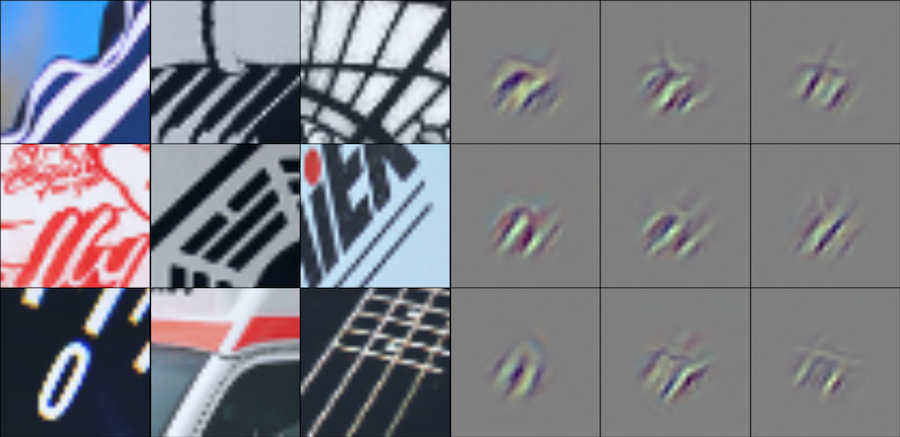}
  \caption{res2b and res2c feature visualizations. Same filters as in figure~\ref{fig:res2a}. }
  \label{fig:res2bc}
\end{figure}

To visualize the activation of this maximal unit, the image is fed through to the unit's layer, all the other activations in that layer are zeroed out, and that layer is then passed backwards using guided backpropagation. When visualizing the top-9 images, we select the patch corresponding to the receptive field of the unit.

\section{Results and Analysis}

In figures~\ref{fig:res2a} and ~\ref{fig:res2bc}, we have visualizations of channels across the 2a, 2b, and 2c building blocks (res2a, res2b, res2c). Filter 12 appears to recognize complex lined patterns, becoming more discriminative in res2b, and slightly more discriminative in res2c. Filter 79 does not appear to change substantially, which is consistent with the intuition that residual blocks are preconditioned to the identity mapping. Filter 150 exhibits refinement: at res2a, it recognizes slightly curved dark lines, but at res2b and res2c begins to recognize loops and tighter curves. Filter 210 does not appear to recognize anything in res2a, but shifts to recognize parallel lines in res2b and res2c.

\begin{figure}[h]
  \centering
  \includegraphics[width=8cm]{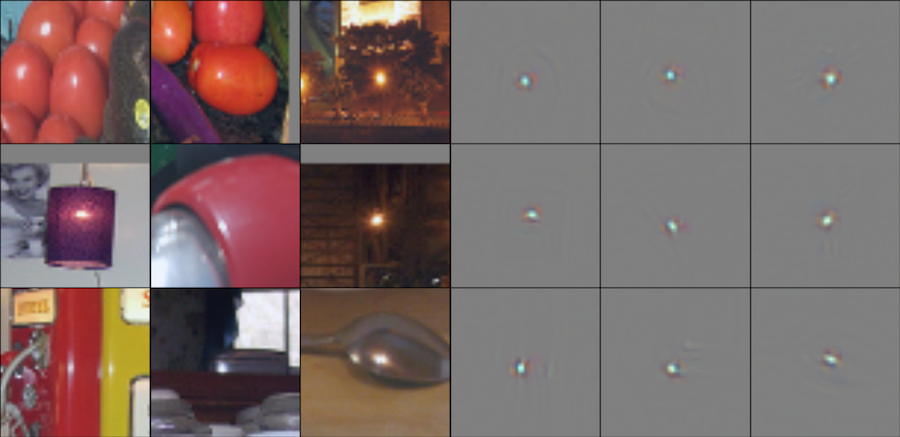}
  \noindent\makebox[\linewidth]{\rule{6.9cm}{0pt}}
  \includegraphics[width=8cm]{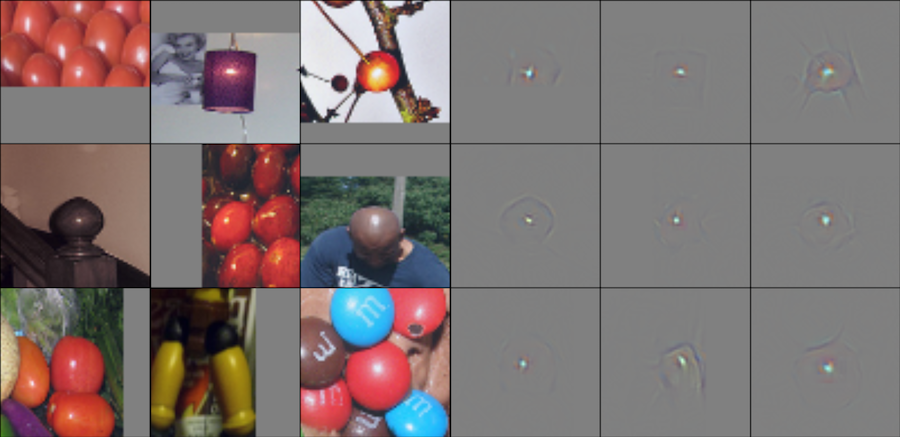}
  \noindent\makebox[\linewidth]{\rule{6.9cm}{0pt}}
  \includegraphics[width=8cm]{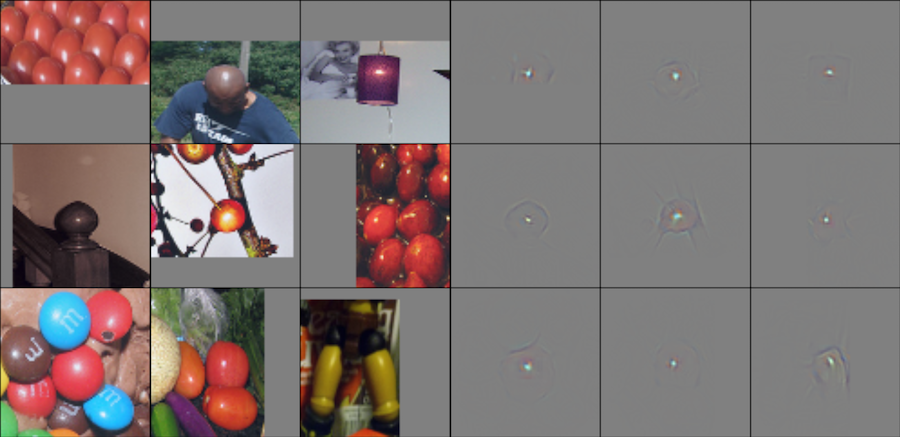}
  \caption{Channel 400, visualized across res3a, res3b, and res3c (left to right, top to bottom). Not randomly chosen.}
   \label{fig:res3abc}
\end{figure}

In figure~\ref{fig:res3abc}, we selected an interesting example of refinement of features. In res3a, it recognizes a single spot of light. In res3b, it is refined to recognize the spot of light with a surrounding context, and further in res3c.

\begin{figure}[H]
  \centering
  \includegraphics[width=6.5cm]{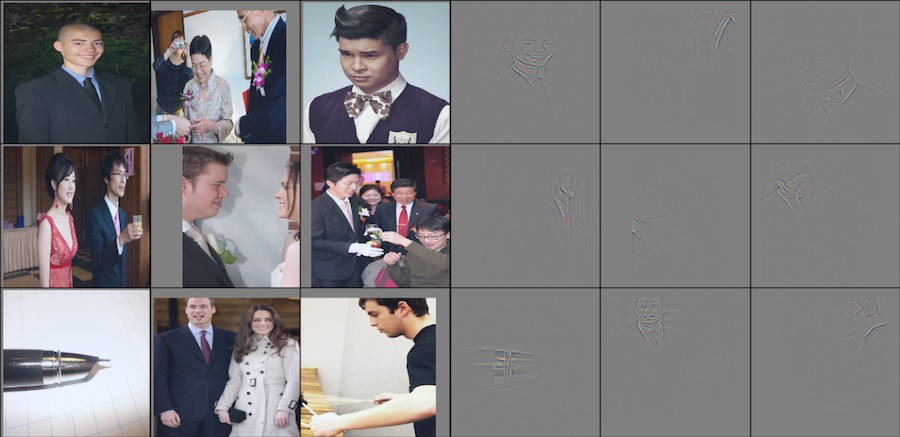}
  \includegraphics[width=6.5cm]{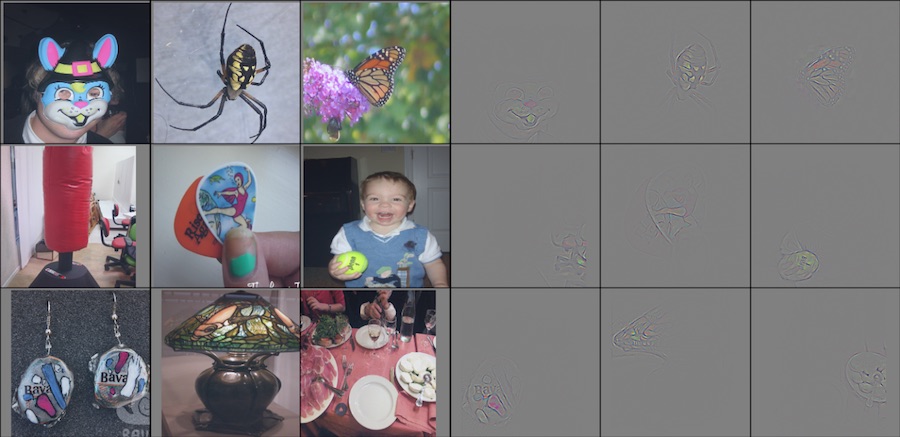}
  
  \smallskip
  \includegraphics[width=6.5cm]{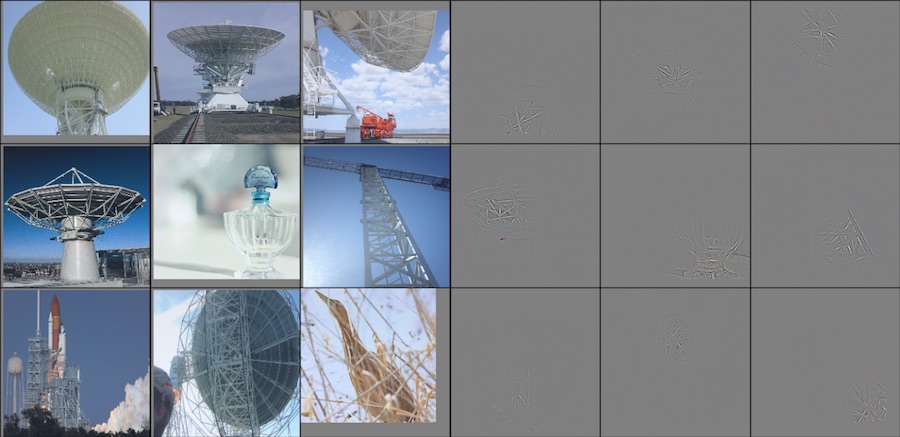}
  \includegraphics[width=6.5cm]{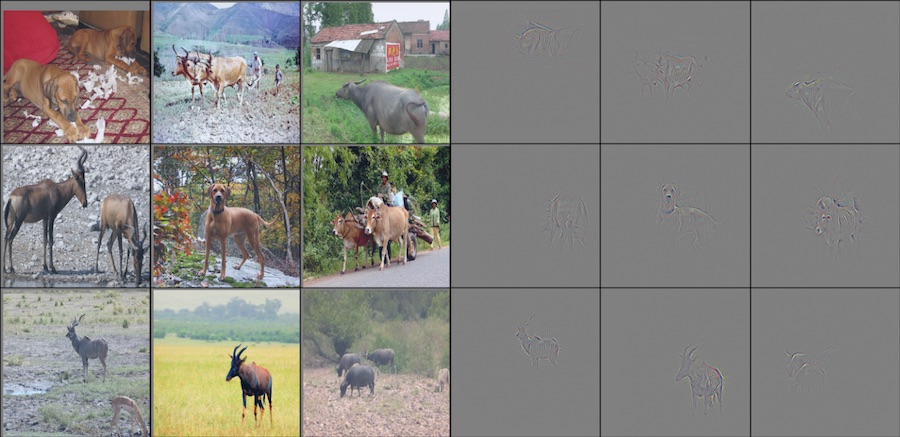}
  \caption{res5a visualizations (random). From left to right, top to bottom: filters 7, 149, 1068, and 1620. Zoom to see deconvolution details.}
  \label{fig:res5a}

  \smallskip
  \includegraphics[width=6.5cm]{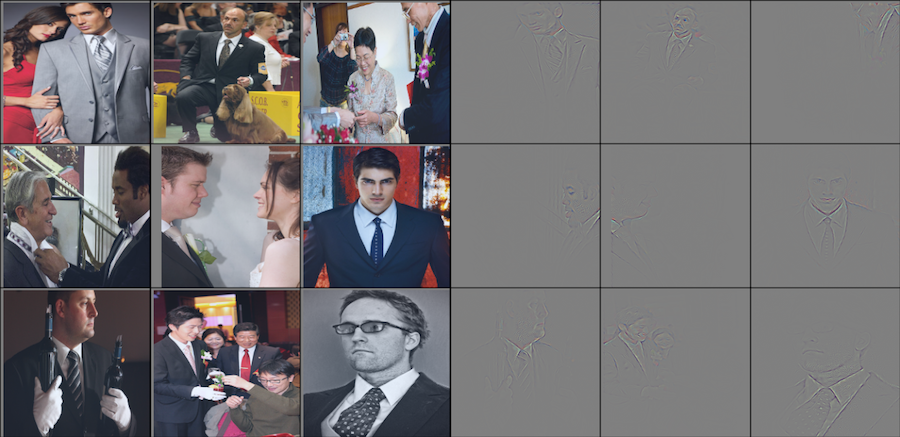}
  \includegraphics[width=6.5cm]{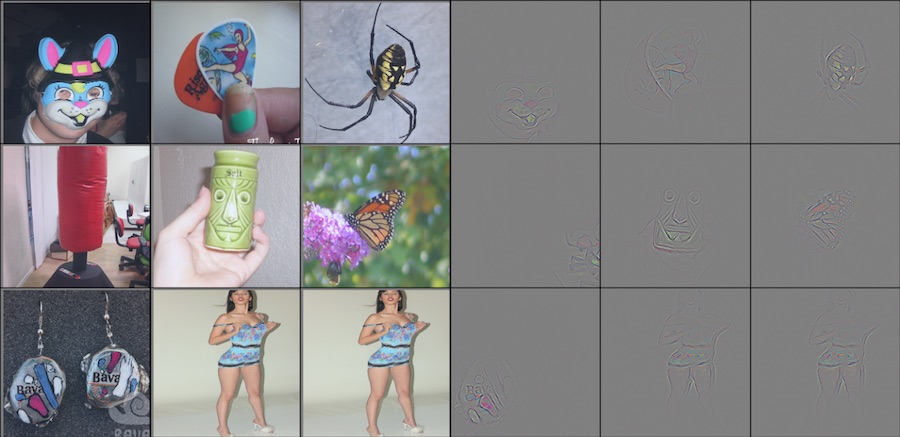}
  \includegraphics[width=6.5cm]{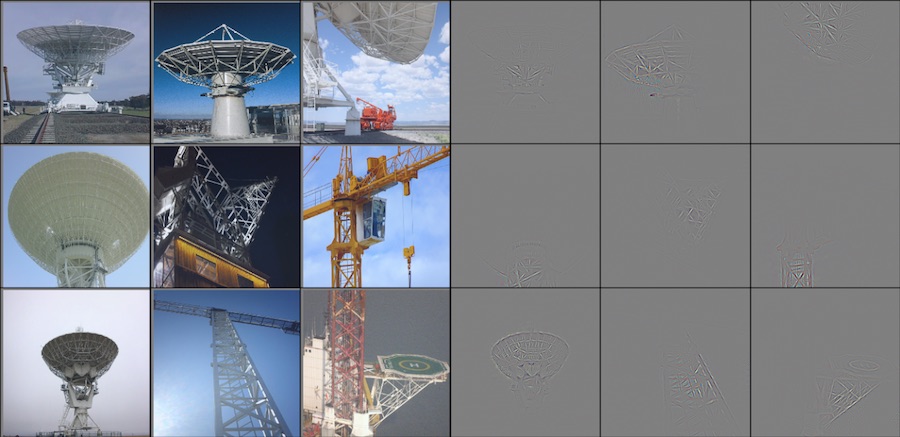}
  \includegraphics[width=6.5cm]{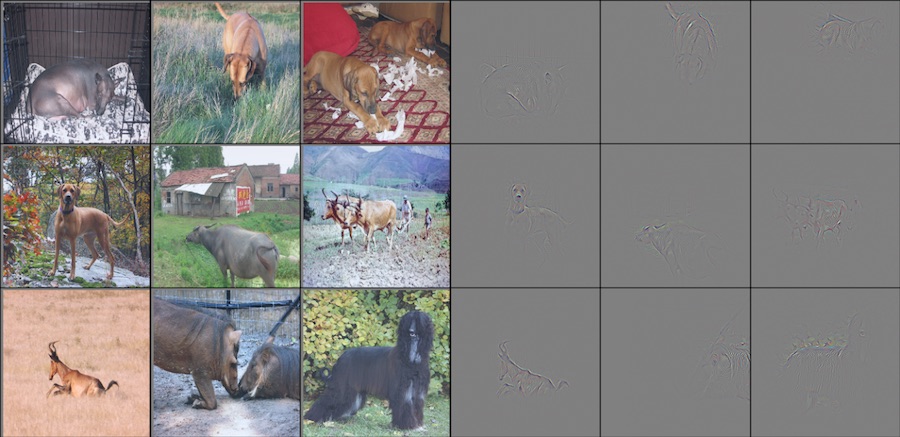}\\
  
  \includegraphics[width=6.5cm]{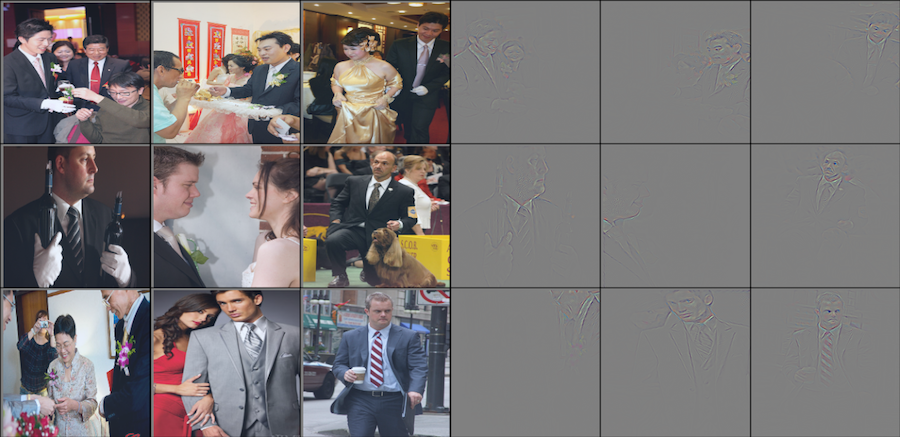}
  \includegraphics[width=6.5cm]{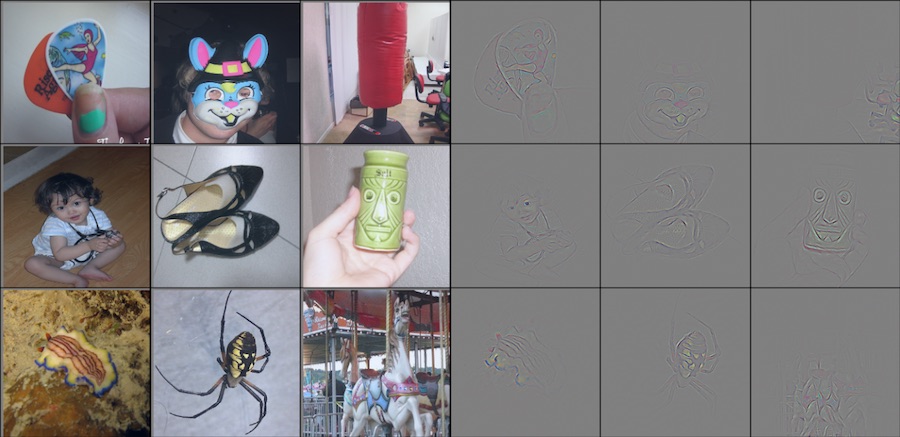}
  \includegraphics[width=6.5cm]{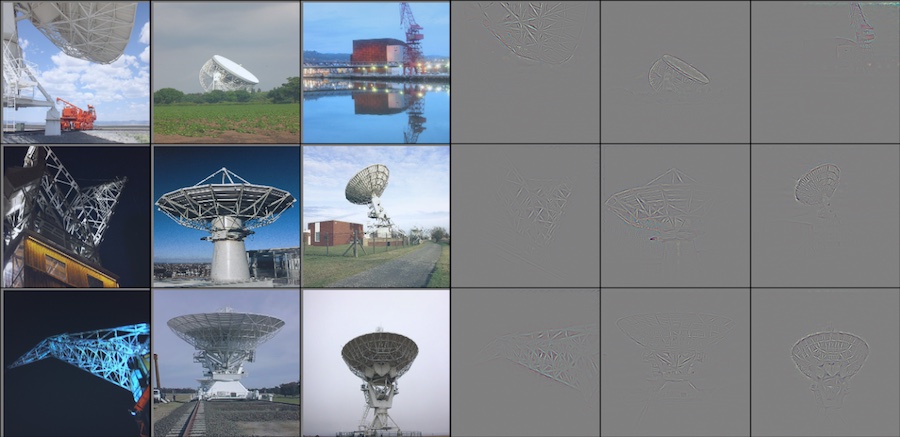}
  \includegraphics[width=6.5cm]{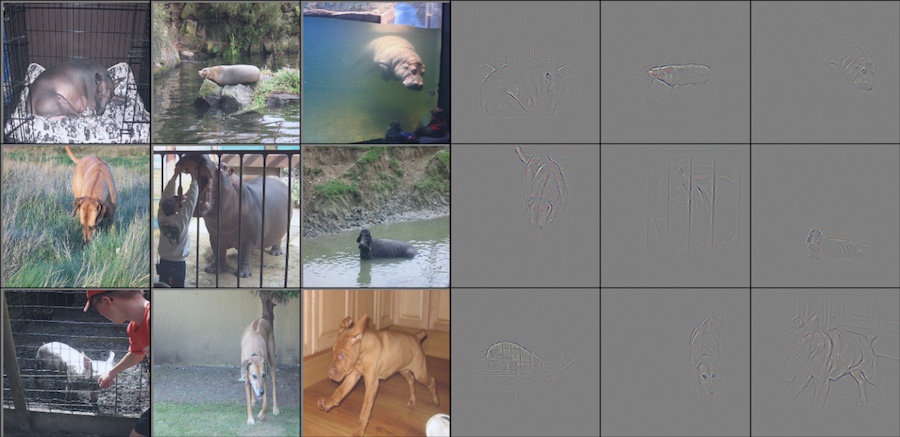}

  \caption{res5b and res5c features. Same filters as in figure~\ref{fig:res5a}. Zoom to see deconvolution details. }
  \label{fig:res5bc}
\end{figure}

In figures~\ref{fig:res5a} and ~\ref{fig:res5bc}  we randomly chose 4 different filters to visualize. All exhibit the refinement effect. Filter 7 appears to recognize men in suits: initially 2 out of the top-9 activations are not of men in suits, but at res5c all are. Examining the deconvolution visualization confirms that this feature is not necessarily focusing on recognizing weddings: it appears the unit fixates on suit collars. Filter 149 recognizes a mosaic pattern, with little apparent refinement. Filter 1068 appears to recognize truss structures, initially containing a perfume bottle in its top activations, before later being dominated by satellite dishes and cranes. Filter 1620 initially highly activates on horned animals, with these activations being refined away by res5c.

\begin{figure}[h]

  \includegraphics[width=15cm]{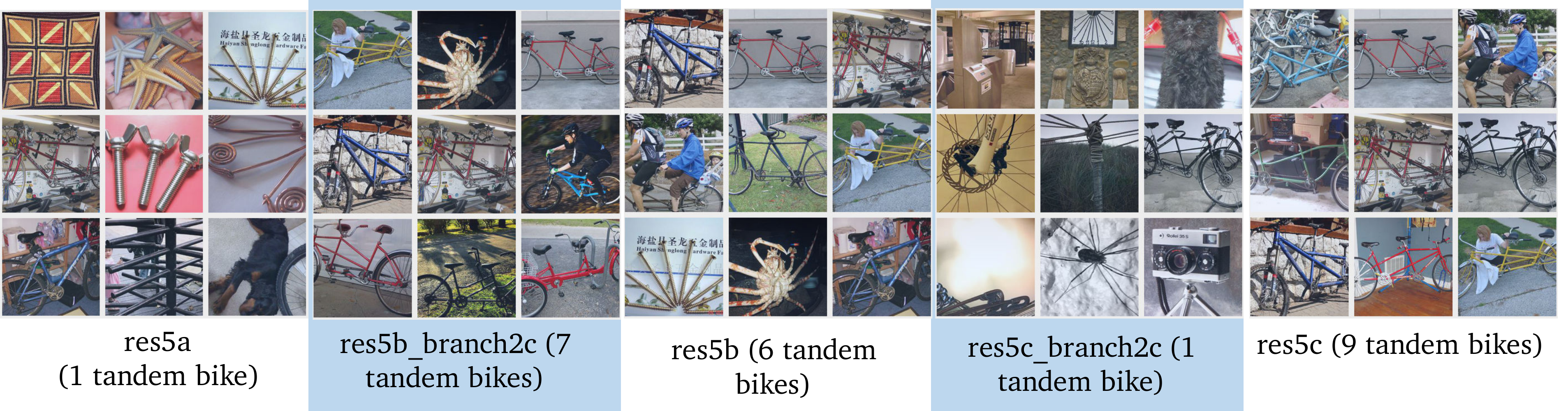}
  \caption{The top-9 activating images of channel 1660 in layers res5a, res5b, and res5c (along with intermediate layers res5b\_branch2c and res5c\_branch2c). The activation on tandem bikes is successively increased. }
  \label{fig:tandem}
\end{figure}

In figure~\ref{fig:tandem}, we also selected another interesting example of refinement of object recognition. Channel 1660 in res5a has 1 tandem bicycle in its top-9 activations. In res5b, the same channel is maximally activated by 6 tandem bicycles, and by the final residual block of that feature size, res5c, all of the top-9 inputs are tandem bicycles. We speculate that the filters in the branches (e.g. res5b\_branch2c and res5c\_branch2c), focus on specific sub-features that help improve the detection of the feature. We see that res5b\_branch2c (channel 1660) latches onto bikes with two seats and res5c\_branch2c (channel 1660) activates on the spokes.

\section{Additional Results}

To see the benefits of ResNet's architecture, we briefly examined AlexNet. AlexNet does not have these identity mappings, so to find units on different layers that correspond we compared the top-9 activation inputs from AlexNet conv4 to those of conv5. The units which matched on more than 1 input in common were few, and there were few examples of refinement as clear as in the residual network.

\begin{figure}[h]
  \centering
  \includegraphics[width=6cm]{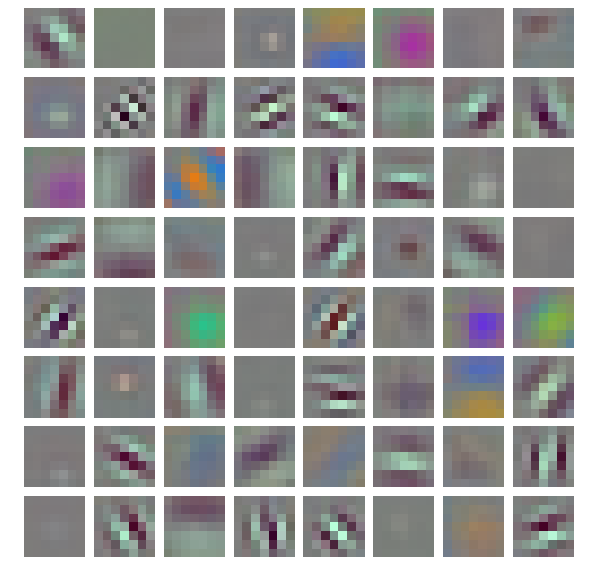}
  \caption{Pixel map of conv1 kernel values.}
  \label{fig:conv1}
\end{figure}

In figure~\ref{fig:conv1} we also directly visualized the kernels in conv1 and found they were similar to AlexNet.

\section{Future Work}

Several ideas came up during the course of this project that were not pursued due to time and compute constraints. These present interesting avenues for exploration:

\begin{enumerate}
\item We did not examine the 1x1, 3x3, and 1x1 convolutions inside each residual block (e.g. inside res2a). These are particularly interesting because the first 1x1 convolution downsamples the channels, the 3x3 convolution maintains the channel dimensionality, and the final 1x1 convolution upsamples (e.g. 1024 channels $\rightarrow$ 256 channels  $\rightarrow$ 256 channels  $\rightarrow$ 1024 channels). It would be interesting to visualize these branched, additive, features.

\item Across res* boundaries (e.g. res3c $\rightarrow$ res4a) the number of channels doubles and a projection building block is used (e.g. res4a). We did not visualize differences across these boundaries because there are no obvious correspondences between filters (unlike with the residual shortcuts). In other words, channel 60 in res4f does not map to channel 60 in res5a It would be interesting to find those correspondences and visualize the evolution of higher-level concepts. Some preliminary exploration found that a feature in the previous layer (e.g. res4f) would usually have more than one corresponding feature (sharing multiple top-9 activation patches) in the next layer (e.g. res5a), and those corresponding features would subsequently diverge.

\item In the past several months, He et al. have made several improvements to the architecture of residual networks \cite{He16}. We could qualitatively evaluate those improvements.

\item Having seen the refinement effect of these residual shortcuts, we could try exploiting this property to expand existing architectures. We speculate it would be possible to take a pretrained 7-layer AlexNet architecture, insert 0-initialized residual blocks of the same channel dimensionality before and after each pretrained layer, and then resume training.
\end{enumerate}

\subsubsection*{Acknowledgments}

Thanks to the CS 280 course staff: Alyosha, Trevor, Deepak, Jeff, and Richard. Thanks Alyosha and Deepak for comments during the poster session.

\begin{figure}[h]
\centering
What is Cory Hall?

\includegraphics[width=4cm]{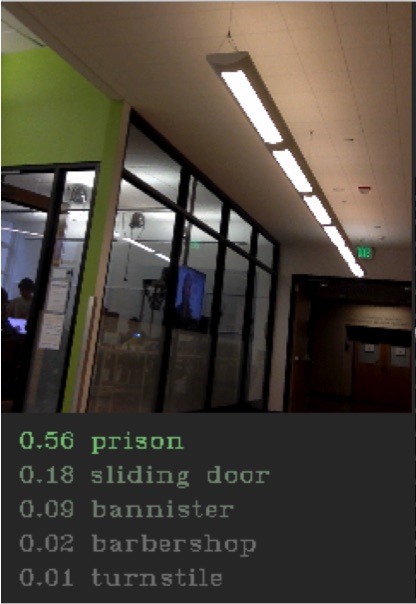}
\end{figure}

\small
\bibliographystyle{plain}
\bibliography{main}
\end{document}